# Special and General Computational Aesthetics

Ganqing Luo[1,2], Chuan Li[1], Liaoruxing Zhang[3,5], Xin Jin[4,5,*]

[1] Sichuan Fine Arts Institute, Chongqing 401331, China
[2] Chongqing College of Artificial Intelligence, Chongqing 401336, China
[3] Beihang University, Beijing 100191, China
[4] Beijing Electronic Science and Technology Institute, Beijing 100070, China
[5] State Key Laboratory of General Artificial Intelligence,
Beijing Institute for General Artificial Intelligence (BIGAI), Beijing 100080, China
* Corresponding author: jinxin@besti.edu.cn

## Abstract

Computational aesthetics is a field at the intersection of computer science, artificial intelligence, aesthetics, and other disciplines. As AI has developed, the field has formed relations of inclusion, intersection, or support with computational art, AI art, AI aesthetics, AI-generated content (AIGC), computational creativity, affective computing, and neuroaesthetics. Without distinguishing scope, objectives, and methods, it may be narrowed to image aesthetic quality assessment or expanded into an umbrella term for AI's involvement in art.

We analyze its internal structure and its relations with neighboring concepts, and propose a two-level scope: special computational aesthetics (SCA) and general computational aesthetics (GCA). SCA centers on aesthetic measurement and aesthetic process modeling. The former primarily corresponds to "Aesthetic Perception": computationally identifying and representing aesthetic attributes and predicting human aesthetic evaluations; the latter primarily to "Aesthetic Understanding": computationally describing, testing, and explaining how human aesthetic perception and judgment are formed. GCA extends this core to include "Aesthetic Generation," namely artistic generation guided by computational aesthetic models. Some computational aesthetic tasks can also use non-AI methods. Research in this scope using AI algorithms and models we define as AI aesthetics, which may involve the special core and the general extension. "Aesthetic Generation" intersects with AIGC, AI art, and computational creativity, but belongs to the general extension defined here only when computational aesthetic models substantively constrain, guide, or optimize artwork generation. We also draw on Feng Peng's discussion of the computability of beauty, aesthetic experience, and art as a theoretical reference for clarifying boundaries between computational aesthetics and related concepts.

Keywords: computational aesthetics; aesthetic measurement; aesthetic process modeling; aesthetic generation

## 1 Special and General Computational Aesthetics

According to the account cited by Denzler et al., the Encyclopaedia Britannica defines computational aesthetics as a branch of artificial intelligence concerned with the computational evaluation of beauty in human creative expressions such as music, visual art, poetry, and chess endgame studies. It also notes that such research may enhance our understanding of human aesthetic perception.[1] In his introduction to the lecture “Computational Aesthetics,” Xin Jin summarizes the relevant capabilities as “Aesthetic Perception,” “Aesthetic Understanding,” and “Aesthetic Generation.”[2] Building on these questions and on subsequent research, we place Aesthetic Perception and Aesthetic Understanding within special computational aesthetics (Special Computational Aesthetics, SCA) and treat artistic generation guided by computational aesthetic models as an extension of general computational aesthetics (General Computational Aesthetics, GCA).

We do not restrict computational aesthetics to research that uses artificial intelligence. Computational aesthetics also includes non-AI research based on mathematical formulas and formal rules. The portion that uses AI algorithms and models is termed AI aesthetics in the sense of computational research adopted here.

Viewed in terms of disciplinary genealogy, formal and quantitative inquiries into aesthetic questions predate contemporary machine learning applications. Birkhoff raised the question of aesthetic measurement in 1928 and, in Aesthetic Measure (1933), systematically discussed measuring aesthetic value through the relation between order and complexity.[3] In the 1950s and 1960s, Moles and Bense advanced information aesthetics, which analyzes artistic form through information, redundancy, and complexity.[4] In 1993, Scha and Bod published the Dutch-language paper “Computationele Esthetica” (“Computational Aesthetics”), an early instance of the term's appearance in research discourse.[5]

Since the beginning of the twenty-first century, advances in computer vision and machine learning have supported computational modeling of the relationships between features of works and human aesthetic evaluations. With the development of deep learning, large-scale aesthetic datasets, and multimodal models, image aesthetic assessment has expanded to encompass aesthetic classification, aesthetic rating, aesthetic distribution prediction, and aesthetic attribute analysis, as well as aesthetic commentary, aesthetic question answering, and aesthetic guidance; large-scale aesthetic models and agents in turn combine multiple tasks within a single system. Meanwhile, theoretical and experimental work in experimental psychology, cognitive science, and neuroaesthetics has provided a basis for research that uses behavioral, physiological, and neural data to examine how human aesthetic perception and judgment are formed. Computational aesthetic models have also increasingly been used to guide artistic generation.

The expansion of the research scope has made questions about the internal structure of computational aesthetics and about its boundaries with neighboring concepts more prominent. Aesthetic measurement tasks—such as aesthetic attribute recognition, score prediction, and multimodal assessment—together with aesthetic process modeling, form the special core of computational aesthetics, although the two differ in their research objectives. Computational aesthetic models can also guide artistic generation through evaluation and feedback, supporting practices in AI-generated content (AIGC), AI art, and computational creativity. These relationships therefore need to be clarified in terms of research scope, task objectives, and technical methods.

Existing research has already incorporated generative questions into computational aesthetics. The review by Bo et al. is organized primarily around two threads—aesthetic measurement and quantification, and generative art—and proposes a design framework that combines evaluation and generation.[6] We are the first to propose a hierarchical distinction between special and general computational aesthetics: within the special core, we distinguish aesthetic measurement from aesthetic process modeling and place artistic generation substantively guided by computational aesthetic models within the general extension. Here “special/general” divides research scopes, the triad “Aesthetic Perception / Aesthetic Understanding / Aesthetic Generation” summarizes the principal research questions, and “whether AI is used” distinguishes technical methods.

We adopt conceptual analysis and comparison of representative literature to clarify these concepts along four dimensions: research objects, research objectives, methods, and evidence. The analysis begins by taking Feng Peng's discussion of the computability of beauty, aesthetic experience, and art as a theoretical reference; it then reviews aesthetic measurement and its extension into further tasks, compares computational aesthetics with aesthetic computing, AI aesthetics, and other neighboring concepts, and finally discusses the evidential requirements of aesthetic process modeling and the contexts in which cross-language terminology is used.

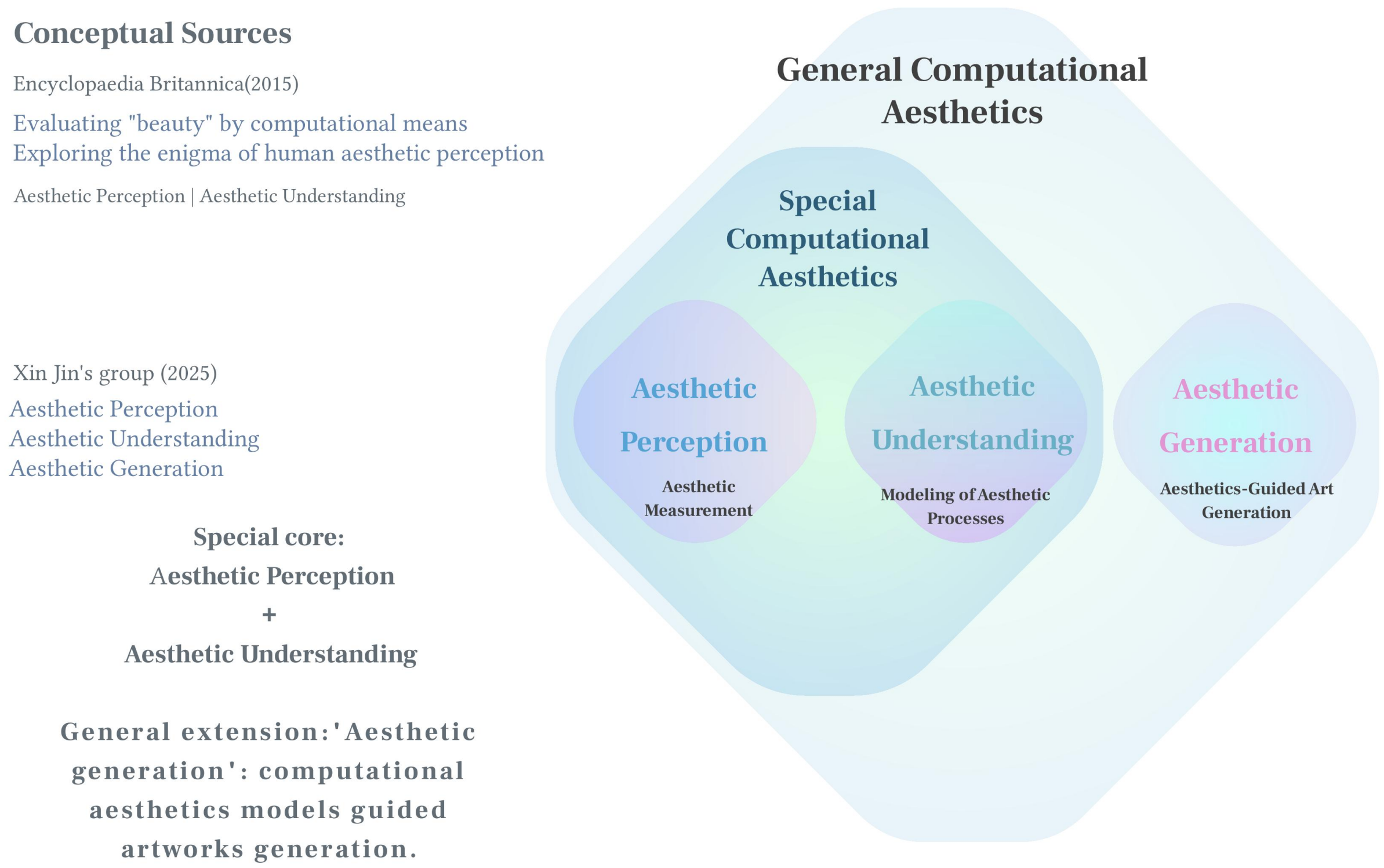

Figure 1. The Hierarchical Structure of Special and General Computational Aesthetics and Their Conceptual Sources. Drawn by the authors based on the relevant definition in the Encyclopaedia Britannica as cited by Denzler et al. (2016), and the Baidu Baike entry "Computational Aesthetics." The special/general hierarchy is our analysis.

## 2 The Computability of Beauty, Aesthetic Experience, and Art

The scope of computational aesthetics concerns both the use of computational methods and how beauty, aesthetic experience, and art can become objects of computational research. In "How Is Computational Aesthetics Possible?", Feng Peng examines the conditions for establishing computational aesthetics, starting with the objects of aesthetic research. He notes that although their definition has long been disputed, beauty, aesthetic experience, and art remain three basic objects commonly accepted by researchers, and examines the computability of each. We take his discussion as a theoretical reference for the computability of aesthetic questions.[7]

### 2.1 The computability of beauty

Feng Peng starts with the recognizability of beauty to discuss the possibility of studying it computationally. He notes that beauty may be difficult to express accurately, but inexpressibility does not imply unrecognizability. To define the scope of his discussion, he primarily restricts "beauty" to style or features, temporarily setting aside its other meanings as an evaluative or ontological concept.

The computability of beauty also concerns observable regularities shared across aesthetic judgments. Feng Peng cites Birkhoff's formal measure and the measured relationships that Shafer et al. established between visual variables in landscape photographs and participants' preferences to discuss how some judgments relate to describable factors. He also uses biological rules, cultural habits, and individual strategies to explain the factors behind differences in judgment and their respective scopes.[7]

These discussions address the recognition of aesthetic features, shared regularities in aesthetic judgment, and the factors behind differences in judgment, providing a reference for the research questions of aesthetic measurement.

### 2.2 The computability of aesthetic experience

In discussing the computability of "aesthetic appreciation," Feng Peng first distinguishes "beauty" from "aesthetic appreciation." In this section, aesthetic appreciation primarily means aesthetic experience. It includes the appreciation of beauty, the sublime, tragedy, comedy, the comic, absurdity, ugliness, and other aesthetic categories. Aesthetic experience may also involve complex emotions, such as the pain evoked by the sublime or tragedy. Drawing on Aristotle's discussion of tragic "catharsis," Burke's analysis of the sublime, and Dubos's account of "harmless passions," Feng Peng notes that pleasure may, in turn, compensate for this pain.

Feng Peng thus proposes that "the computability of aesthetic experience is, in fact, the computability of emotion."[7] His discussion extends from pleasure to joy and sadness, hope and fear, ecstasy and despair, and concerns aesthetic categories formed by emotional types and artistic styles. With developments in cognitive psychology, neuroaesthetics, brain science, and digital humanities, emotional types can be studied using measurement indicators. Words describing aesthetic emotions and artistic styles can also be collected and analyzed statistically. Categories such as gracefulness, somberness, and ethereality express both artistic styles and emotional feelings. The beautiful and the sublime, and tragedy and comedy, contrast with one another, yet may all express positive aesthetic value. Organizing emotional types, artistic styles, and their relationships offers a possible way to study aesthetic experience computationally.

### 2.3 The computability of art

For the computability of "art," Feng Peng focuses on creativity. The difficulty of finding a stable, closed definition of art is closely related to the continual emergence of originality, novelty, and innovation in art. If creativity is an important feature of art, then the computability of art also concerns whether creation can be explained and computed.

Feng Peng considers decomposing the creative process into steps or factors a possible way to examine its computability. He cites stages such as preparation, incubation, inspiration, and production, while noting that inspiration remains difficult to explain fully. Analyzing these stages and their changes makes further computation and explanation of the creative process possible.

Feng Peng takes a more cautious view of AI creation. He argues that computational agents may follow cognitive and emotional patterns different from human ones. People currently tend to evaluate AI "creation" through its results, while its internal processes remain a black box. He therefore holds that AI creative processes cannot yet be explained through human mechanisms of creation.[7]

Feng Peng's discussion offers a philosophical analysis of the computability of aesthetic questions. The recognizability of beauty, shared regularities, and factors behind differences provide references for aesthetic measurement; emotional questions in aesthetic experience relate to affective computing; and artistic creativity concerns computational creativity. We connect these questions to computational research at the level of research questions, not through one-to-one correspondences with technical tasks. Peng's landscape-preference examples involve evaluation practice, but his article does not systematically review the task system of modern image aesthetic assessment.

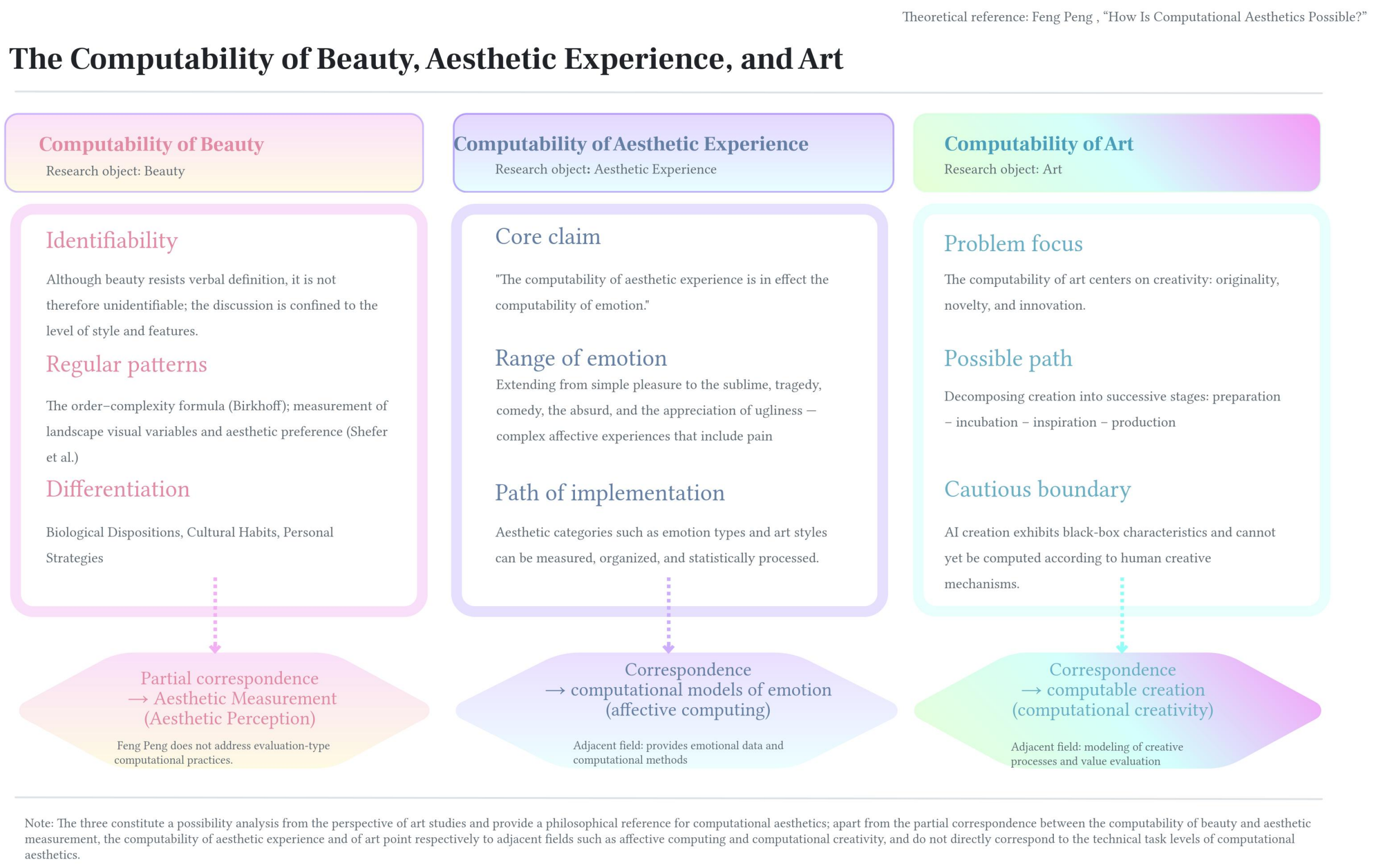

Figure 2. The Computability of Beauty, Aesthetic Experience, and Art and Their Connections with Computational Research. Compiled and drawn by the authors based on Feng Peng's (2025) discussion and our framework. The correspondences between fields are partial connections we draw. For aesthetic assessment, we distinguish Peng's examples of rating calculations from modern assessment-task systems.

# 3 Aesthetic Measurement: From Formal Quantification to Multimodal Aesthetic Assessment

## 3.1 The conceptual boundaries of aesthetic measurement: from artistic quantification to aesthetic assessment

In this discussion, aesthetic measurement primarily corresponds to "Aesthetic Perception" and concerns the computational representation and measurement of the attributes of aesthetic objects, aesthetic evaluations, and their relationships.

Perc reviews quantitative research on artistic expression from the perspectives of physics and network science, and identifies quantifying beauty and modeling human aesthetic perception as important directions in computational aesthetics. His review covers structure, complexity, entropy, and evolutionary relationships in culinary, visual, musical, and literary arts, while distinguishing the quantification of artistic expression from the direct measurement of beauty itself.[8] Computing an image's entropy, recognizing a style, or statistically analyzing musical structure can reveal features of a work. To determine whether these analyses constitute aesthetic measurement, one must also explain how the selected variables represent aesthetic attributes or establish testable relationships with aesthetic evaluations.

## 3.2 The development of aesthetic measurement: from formal relationships to evaluation models

Aesthetic measurement is a research direction that emerged relatively early within computational aesthetics and has a relatively mature task system.

Early mathematical aesthetics sought to connect formal factors with aesthetic value. Birkhoff proposed $M = O/C$, where M is aesthetic measure, O is order, and C is complexity.[3] The formula makes order and complexity computable variables, although its applicability still needs to be tested against specific objects and human evaluations.

Information aesthetics describes the organization of artistic forms through variables such as information, redundancy, and complexity, providing quantitative methods for analyzing the structure of works.[4]

Compared with early studies of formal regularities, data-driven image aesthetic assessment puts greater emphasis on training and validating models with annotated data and studying statistical associations between image features and human aesthetic judgments.

Tong et al.'s 2004 study began by classifying photographs taken by professional photographers or home users. The authors trained classifiers using low-level features such as color, texture, shape, and energy, and mapped the classification outputs to an overall quality indicator that could be compared with human ratings.[9] Figure 3 places this work among the early studies of image aesthetic classification because it explored the overall evaluation of photographic images. The original study's direct labels, however, identify photographer provenance, not human-annotated beauty and ugliness. Provenance labels and direct aesthetic ratings differ in meaning, and model results should be interpreted according to the labels used.

Handcrafted-feature methods rely on visual factors that researchers design in advance, so photographic experience, formal rules, and feature selection influence these methods.

Image content also affects feature selection. Light and shadow in portrait photography, spatial layers in landscape photography, and structural proportions in architectural photography, for example, may be attributes of interest in different tasks. Evaluation models therefore need to consider how image content relates to the selected attributes.

The AVA dataset, introduced in 2012, provides a large number of images with human ratings, forming a basis for training and comparing large-scale evaluation models.[10] In their 2014 RAPID study, Lu et al. applied deep convolutional networks to image aesthetic assessment, advancing the learning of multilayer visual representations.[11] Deep learning reduces some reliance on handcrafted features, although data selection, label definitions, and the evaluating population still affect model performance.

### 3.3 The expansion of aesthetic measurement tasks: from score prediction to multimodal aesthetic assessment

Image aesthetic assessment tasks have gradually expanded from overall classification and scoring to rating distributions, attribute analysis, language-based critique, question answering, and guidance.

In their 2018 review, Xin Jin et al. summarized five task levels in image aesthetic assessment: aesthetic classification, aesthetic rating, aesthetic distribution, aesthetic factors (aesthetic attributes), and aesthetic description.[12] In 2022, the team proposed image aesthetic visual question answering and aesthetic language guidance as separate tasks. [13, 14] Figure 3 also identifies representative achievements of the Computational Aesthetics Lab (CALab) at Beijing Electronic Science and Technology Institute and related research teams. Following Figure 3, we organize these tasks into a seven-level task model: aesthetic classification, aesthetic rating, aesthetic distribution, aesthetic attributes, aesthetic commentary, aesthetic question answering, and aesthetic guidance. The model describes a functional expansion from representing evaluations and communicating through language to providing suggestions for improvement.

Figure 3 groups the tasks by their main functions and outputs. Classification, scoring, distribution, and attribute analysis fall under numerical assessment; commentary and question answering fall under linguistic assessment, with “art criticism” and “linguistic communication” summarizing their respective functions. Aesthetic guidance, which provides suggestions for adjustment, is listed separately. Numerical assessment also includes structured outputs such as category labels and distributions, and aesthetic guidance can take a linguistic form.

Aesthetic classification concerns categorical judgments of an image's overall aesthetic quality, aesthetic rating expresses evaluation as a continuous numerical value, and aesthetic distributions describe agreement and disagreement in group judgments. The same mean score may come from highly consistent ratings or sharply polarized evaluations by different individuals. Rating distributions can therefore provide information beyond the mean.

Aesthetic attribute tasks concern evaluation-related factors such as composition, color, light and shadow, depth of field, sharpness and blur, motion, and interest. Aesthetic commentary (called aesthetic description in the earlier review), aesthetic question answering, and aesthetic guidance expand how evaluation is expressed. They use language to describe attributes and evaluative reasons and provide corresponding suggestions for improvement. Figure 3 uses “art criticism” to summarize the expressive function of aesthetic commentary. Here, the term means critique tasks concerning aesthetic attributes and evaluative reasons, rather than the full theory and practice of art criticism.

AesExpert aims to develop a multimodal aesthetic perception model that combines image analysis with linguistic responses to support analysis and communication about image aesthetic attributes.[15] The reasons generated by the model may help users understand its evaluations, but whether these reasons faithfully reflect its computational process and whether they correspond to human mechanisms of aesthetic judgment require separate validation.

When aesthetic guidance takes the form of advice, it remains an extended task of aesthetic measurement. It connects to the aesthetic generation extension only when recommendations or evaluations constrain the generation, editing, and optimization of artworks. Generating evaluative text does not itself constitute artistic generation in the sense used here. In the technical literature, "multimodal aesthetic understanding" may refer to attribute analysis, semantic processing, or linguistic response; the word "understanding" alone does not justify classifying such work as aesthetic process modeling as defined here.

### 3.4 The expansion of evaluation objects, standards, and applications

Aesthetic measurement has expanded from photographic images to other media. APDD organizes aesthetic attributes and evaluations by categories of paintings and drawings; [16] VADB provides overall scores, attribute scores, language comments, and other annotations for video; [17] and MADB organizes overall scores, perceptual dimensions, and written comments for music.[18] Cross-media evaluation requires tasks and annotations designed for each medium's temporal organization, structural features, and context of appreciation.

The organization of evaluation information has also entered standardization practice. ITU-T F.740.4 specifies metadata for image aesthetics assessment with aesthetic attributes in mobile terminal computational photography systems. It organizes the attributes around the photographer, camera, and audience, and regulates how aesthetic assessment metadata are described and organized.[19]

The Huawei Pura 80 series uses AI-assisted composition to provide shooting recommendations.[20] Research systems also use agents to organize multistep actions. Yao et al.'s PhotoAgent supports autonomous photo editing through aesthetic planning, visual feedback, and tool execution.[21] Che et al.'s PhotoAgent, which has the same name, connects aesthetic objectives with geometric constraints for robotic photography.[22] These studies bring evaluations or recommendations into editing and capture workflows, respectively. Whether they involve aesthetic generation as defined here also depends on the artistic context and the aesthetic model's specific role.

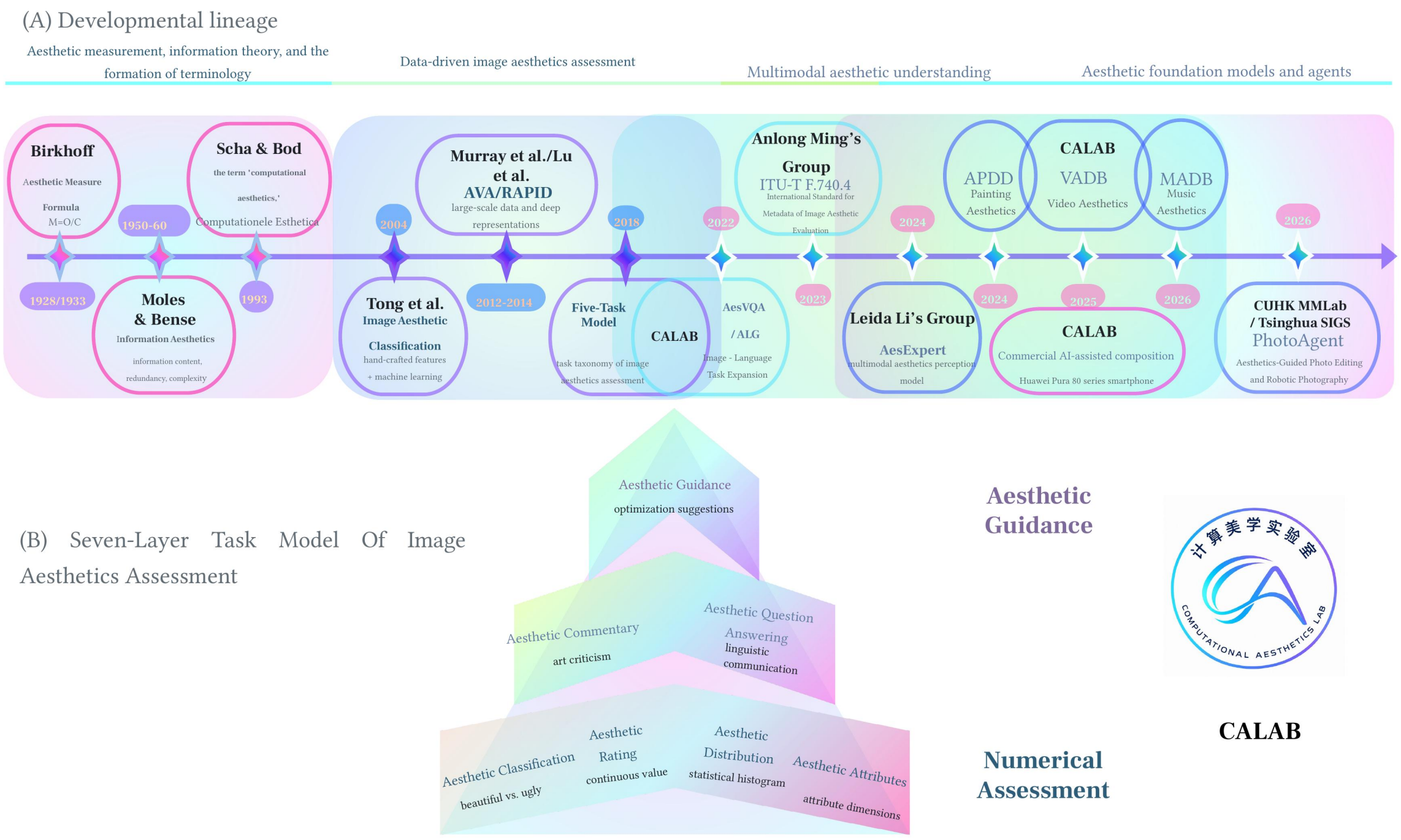

Figure 3. The Development of Computational Aesthetics and the Expansion of Aesthetic Measurement Tasks. Compiled from Birkhoff (1933), Scha and Bod (1993), Tong et al. (2004), Murray et al. (2012), Lu et al. (2014), Xin Jin et al. (2018, 2022), ITU-T F.740.4 (2023), Huang et al. (2024), and the materials on APDD, VADB, MADB, Huawei Pura 80, and the two PhotoAgent studies cited in the text. The seven task levels are a synthesis: 2018 corresponds to the five-level foundation, and question answering and guidance were added in 2022. Tong et al.'s placement under "image aesthetic classification" is our interpretation of the research lineage; the study's direct labels identify photographer provenance. The figure presents representative milestones rather than a complete technical history. Drawn by the authors.

## 4 Computational Aesthetics and Neighboring Concepts: Research Objects, Objectives, and Methodological Boundaries

Computational aesthetics has different types of relationships with neighboring concepts. This section distinguishes the research orientations of computational aesthetics and aesthetic computing, explains AI aesthetics as a methodological subset, and then discusses artistic production, content generation, creative processes, and affective and neural research.

These lines of research may coexist within a system. An artistic generation system, for example, may have modules for generation mechanisms, aesthetic evaluation, user feedback, and the exploration of alternatives. Research objects and objectives determine the questions each module addresses; methods and evidence explain how it addresses them and which conclusions can be supported.

### 4.1 Computational aesthetics and aesthetic computing

Fishwick describes aesthetic computing (Aesthetic Computing) as applying the theory and practice of art and aesthetics to computing. It focuses on how aesthetics can inform computational objects and their representation, interaction, and design. Computational aesthetics (Computational Aesthetics), as we use the term, takes aesthetic attributes, aesthetic evaluations, and the processes through which they are formed as objects of computational research.[23]

Fishwick's computational objects include programs, data structures, and mathematical models; his discussion also covers human–computer interaction and visualization. Aesthetics can be used to analyze symmetry, simplicity, and harmony in computational and mathematical objects and to inform the design of their representations and interactions. He describes the former as analytical applications. In the latter, artistic and aesthetic methods change how computational objects are presented.

"Beauty" in mathematics and algorithms is a direct example. Mathematicians may evaluate formulas, proofs, and solution processes for their simplicity, symmetry, unity, or unexpectedness. Programs and algorithms may also be described as "elegant" for their concise structures and clear logic. In discussing mathematical aesthetics, Fishwick mentions aesthetic pleasure derived from theorems and proofs, along with invariance, symmetry, simplicity, proportion, and harmony. Here, the "beauty of algorithms" means human aesthetic judgments of mathematics, programs, algorithms, and their solution processes.[23]

Figure 4 uses "opposite research objects" to indicate the different research orientations of computational aesthetics and aesthetic computing. Researchers can, for example, design visual representations of an algorithm according to aesthetic principles and use computational models to evaluate the aesthetic effects of those representations. The same project may thus involve both orientations. Aesthetic computing includes the representation, interaction, and design of computational objects; it is not limited to evaluating whether an algorithm is beautiful.

### 4.2 AI aesthetics

We use "AI aesthetics" (AI Aesthetics) in a computational research sense: the portion of computational aesthetics that uses AI algorithms and models to study aesthetic questions. To classify a study, one should first determine whether it belongs to computational aesthetics based on its research question and computational treatment, then determine whether it belongs to AI aesthetics based on its methods. AI aesthetics is defined by method and may involve research on Aesthetic Perception, Aesthetic Understanding, and Aesthetic Generation. It is not a fourth task category parallel to these three.

Aesthetic measurement based on explicit formal rules does not automatically become AI research because it produces numerical outputs. Predicting aesthetic ratings with deep learning can belong to both computational aesthetics and AI aesthetics. AI methods can also be used to propose and test hypotheses about aesthetic processing or to guide artistic generation through aesthetic models. Conversely, using AI only to recognize ordinary objects does not justify classifying a study as AI aesthetics.

AI aesthetics also differs from AI art. The fact that a work is AI-generated does not by itself establish the involvement of aesthetic measurement, aesthetic process modeling, or guidance by aesthetic models. AI aesthetic models may evaluate human-created works or natural landscapes as well as AI art. How a work is generated and which methods a study uses therefore need to be assessed separately.

Existing English-language literature does not always use the term within the scope defined above. Manovich's AI Aesthetics discusses the automation of aesthetic choices and cultural production and its effects on aesthetic diversity. It also lists applications of aesthetic rating and, in "Measuring Diversity," raises the question of formally measuring aesthetic diversity. The scoring and formal measurement concern the computational research discussed here, while the book's coverage of cultural production and its social effects gives it a broader scope.[24]

Manovich and Arielli's Artificial Aesthetics discusses AI applications in aesthetics along two dimensions: object/subject and description/generation. In the first chapter, Arielli distinguishes object analysis, object generation, subject research, and subject simulation. He notes that extracting formal patterns does not automatically explain how they affect human perception, emotion, and cognition. This classification uses different dimensions from our task division: simulating aesthetic judgments can serve evaluation prediction, while explaining how judgments form requires corresponding process hypotheses to be proposed and tested.[25]

Thon and Wilde explicitly place the AI aesthetics of AI image generators within a media-aesthetic framework. They distinguish artistics, which concerns artistic judgment, aisthetics, which concerns perception in general, and aesthetic perception in a specific sense. They also note that the field has no dominant conceptual definition yet.[26] Their focus is thus on the perceptual forms, experiences, and contexts of use of AI media, rather than simply renaming AI art as AI aesthetics.

Boden and Edmonds's From Fingers to Digits: An Artificial Aesthetic studies computer art and its relationship with traditional art, with particular attention to generative art and the role of programs. The subtitle concerns programs, creation, and artistic ideas in computer art; its scope cannot be defined simply by the use of AI methods.[27]

In Chinese-language discussions, Feng Tao treats philosophical aesthetics as the principal method of AI aesthetics, while also discussing its possible involvement in programming and specific technical research.[28] Feng Tao primarily treats AI-related phenomena as objects of theoretical research; we classify research by the use of AI methods to address aesthetic questions.

The methodological definition above organizes the computational research discussed here. Other uses in cultural, media, and philosophical research are best understood through their respective questions and methods; citations of these works should make their specific connections to the scope we define explicit.

### 4.3 Computational art

Computational art concerns how computational processes participate in artistic creation and in the generation and operation of works. Algorithms, programs, rules, data, and interactive systems are computational components of artistic production.

Programmed drawing in the 1960s is an early example of computational art. Boden and Edmonds review the relationship between computational art and generative art, and the roles of different forms of computation in artistic production.[29] In program-generated, fractal, or interactive art, artists design rules and systems through which computational processes participate in the generation and operation of forms.

Computational art primarily concerns the participation of computation in artistic production. Classifying a study as computational aesthetics also requires examining whether it computationally represents and models aesthetic attributes, aesthetic evaluations, or the processes through which they are formed, or substantively guides artistic generation with the relevant models. The fact that a work is generated by a program does not by itself determine whether the study belongs to aesthetics.

### 4.4 AI art and AIGC

Within the generative artistic practices we consider, AI art (AI Art) is a shared subfield of computational art and AI-generated content (AIGC): AI technologies participate in content generation, and the generated content enters practices involving creative intention and an artistic context.

AIGC emphasizes the technical mechanisms through which AI systems generate content. It includes text, images, audio, video, and code, with applications in office documents, commercial design, educational assistance, and other fields.

The literature also discusses AI and art more broadly, including the artistic status of works, human–machine collaboration, the artist's role, and authorship. Cetinic and She's review examines AI research in both art analysis and art creation.[30] These questions can relate to the generative practices considered here, but discussing AI art theoretically is not equivalent to conducting computational aesthetic research with AI methods.

Whether AI-generated content enters artistic practice and whether the related research belongs to computational aesthetics are separate questions. The former concerns creative intention, the presentation of works, and artistic context; the latter depends on aesthetic questions and their computational treatment.

Post hoc analysis of aesthetic attributes or prediction of evaluations for AI-generated artworks can still belong to special computational aesthetics. The aesthetic generation extension of general computational aesthetics is involved only when computational aesthetic models also guide the generation, editing, and optimization of artworks. Figure 4 calls this function "Aesthetics-Guided Art Generation." Within this framework, generation means the generation of artworks, and constraint means that a computational aesthetic model substantively participates in that process. Classification depends on the model's function in creation, rather than who or what generated the work or how the evaluation is expressed.

### 4.5 Computational creativity

Computational creativity studies the computational description, implementation, and evaluation of creativity. It examines the results that systems produce, their behavior and processes, and their environments.

Novelty and value are important aspects of creativity assessment. Jordanous discusses the evaluation of computational creativity from the perspectives of the creative producer, process, product, and environment, showing that evaluation needs to consider specific tasks and contexts.[31] In our account, creative-process modeling and value evaluation are the main questions connecting computational creativity with computational aesthetics.

Differences among generated results do not themselves constitute creativity. Evaluating a generative system requires specifying its task, environment, and criteria for novelty and value, and explaining whether the evidence concerns products or creative processes. Such evaluation need not presuppose the complete reproduction of human psychological processes.

Computational aesthetics can support artistic tasks in computational creativity, although aesthetic assessment and creativity assessment have different objectives. Even when a system generates works that meet aesthetic requirements, its creativity still needs to be examined separately.

The attribute analysis, evaluation, and feedback provided by computational aesthetics can help creative systems explore and select alternatives. Computational creativity's modeling of search, combination, evaluation, and adjustment can also support aesthetics-guided artistic generation. Figure 4 uses “Assistance” to indicate these functional connections.

### 4.6 Affective computing

Affective computing studies computational questions related to, arising from, or influencing emotion, including recognition, modeling, expression, and response.[32] In Figure 4, perceiving, understanding, and generating emotion mean the computational processing of the relevant information and expressions. They do not imply that systems possess human subjective emotional experience.

In art and aesthetic research, affective computing can analyze individuals' emotional responses to works through facial expressions, physiological signals, speech features, or behavioral data.

Feng Peng discusses aesthetic experience through the computability of emotion. Aesthetic process modeling as considered here also needs to examine perception, attention, memory, cultural experience, and value judgment. Recognizing general emotions is insufficient to explain aesthetic experience.

Affective computing therefore provides relevant data and methods for aesthetic process modeling.

### 4.7 Neuroaesthetics and computational neuroaesthetics

Neuroaesthetics studies the brain mechanisms of aesthetic experience and artistic activity from a neuroscientific perspective. It examines people's perceptual, emotional, and cognitive responses to works or aesthetic objects.[33]

Computational neuroaesthetics constructs computational aesthetic models inspired by neuroaesthetics. By connecting stimulus features, behavioral responses, and neural processes, it studies computational relationships in the formation of aesthetic experience. These studies provide neural evidence and testable mechanistic hypotheses for aesthetic process modeling.[34]

These relationships differ in type. AI aesthetics is a subfield of computational aesthetics defined by method. Computational art connects with aesthetic generation when aesthetic models substantively participate in creation. Within the generative practices considered here, AI art is a shared subfield of computational art and AIGC. Computational creativity, affective computing, and neural research can provide relevant models, methods, or evidence; aesthetic computing has a different research orientation.

Table 1 summarizes the concepts' research objects and classification criteria, and Figure 4 presents their relationships.

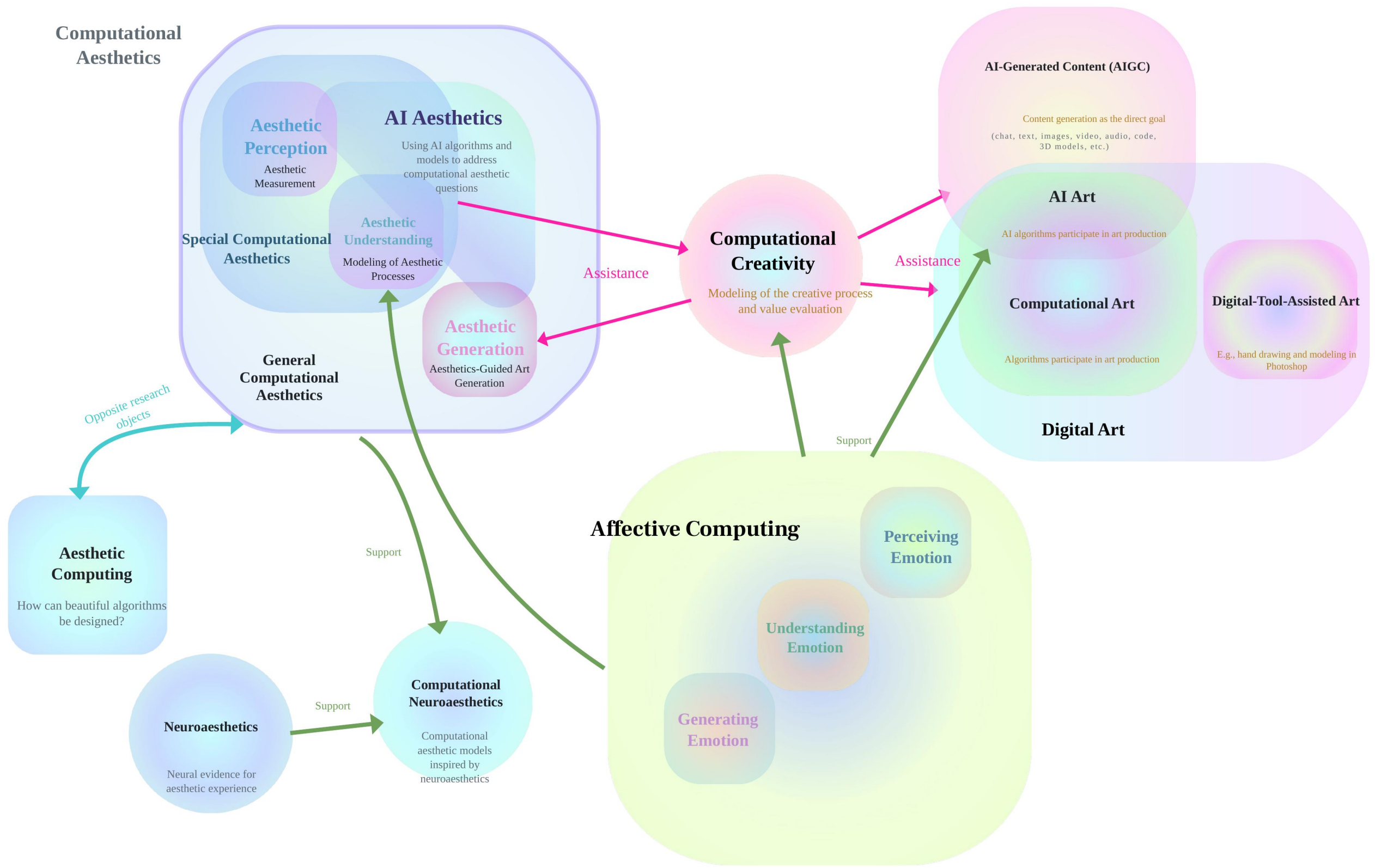

Figure 4. Relationships between Computational Aesthetics and Neighboring Concepts. Compiled and drawn by the authors based on our conceptual distinctions. AI aesthetics is a subfield defined by method; AI art's shared-subfield relationship is limited to the generative practices considered here. “Aesthetics-Guided Art Generation” in the “aesthetic generation” node is limited to artistic generation substantively constrained or guided by computational aesthetic models. Support, Assistance, and opposite research objects indicate methodological and evidential support, functional connections, and differences in research orientation, respectively. The arrows do not indicate identical inclusion relationships among the fields across all questions.

Table 1. Research Objects and Boundaries of Computational Aesthetics and Neighboring Concepts.

| Concept | Direct Research Object | Core Question | Relationship with Computational Aesthetics |
|---|---|---|---|
| Computational Aesthetics | Aesthetic attributes, evaluations, and the processes through which they are formed; artistic generation guided by aesthetic models | How aesthetic questions enter computational representation, prediction, explanation, and generation guidance | The special core comprises aesthetic measurement and aesthetic process modeling; the general scope further includes aesthetic generation |
| AI Aesthetics | Computational aesthetic questions addressed using AI methods | Using AI algorithms and models to address computational aesthetic questions | A subfield defined by method in we; involves some research within the three task types |
| Aesthetic Computing | Computational objects and their representation, interaction, and design | How can beautiful computational methods be designed? | Its research objects are opposite to those of computational aesthetics |
| Digital Art | The use of digital technologies in the creation, production, or presentation of artworks | How can the core characteristics of digital technologies serve as the internal driving force and aesthetic logic of creation? | Computational art is part of digital art |
| Computational Art | Generating artworks through algorithms | How do algorithms participate in the generation and operation of works? | Conditionally intersects with artistic generation guided by aesthetic models |
| AI-Generated Content | AI-generated text, images, video, audio, code, graphics, and other content | How is content generated through AI? | Artworks are only one part of content |
| AI Art | Generating artworks through AI algorithms | How does AI participate in artwork generation? | A shared subfield of computational art and AIGC |
| Computational Creativity | Creative processes, behaviors, products, and evaluation | Creative-process modeling and value evaluation | Aesthetic assessment can support exploration and selection; creativity models can support aesthetic generation |
| Affective Computing | Emotional information and related human–computer interaction | Giving machines the ability to perceive, understand, and generate emotion | Provides relevant data and methods for aesthetic process modeling |
| Neuroaesthetics | Neural mechanisms of artistic creation and aesthetic experience | How does brain activity participate in the relevant experiences and judgments? | Provides neural evidence and mechanistic hypotheses |
| Computational Neuroaesthetics | Computational models based on the neural mechanisms of artistic creation and aesthetic experience | How can computational models represent and test these neural mechanisms? | Supports aesthetic process modeling; is not equivalent to computational aesthetics as a whole |

# 5 Aesthetic Process Modeling: From Evaluation Results to Computational Studies of Formation Mechanisms

## 5.1 From aesthetic measurement to aesthetic process modeling

Aesthetic measurement establishes computational representations or predictive relationships for attributes and evaluations; aesthetic process modeling studies how evaluations are formed. Even when a model assigns high scores and independent data confirm that its scores correlate with human judgments, this remains insufficient to explain how perception, attention, emotion, and memory participate in those judgments.

Aesthetic process modeling needs to express these psychological processes and their interrelationships as testable models. It examines information selection, the roles of emotion and cognition, and how individual experience and cultural context affect evaluation.

Specific studies help clarify this distinction. Drawing on the relationship between convolutional networks and visual processing, Denzler et al. compare network responses to artistic and non-artistic images and examine image statistics such as sparsity and self-similarity. They explore whether networks can be computational models of human aesthetic perception.[1] This offers a way to propose process hypotheses, but similarities between network architecture and the biological visual system are insufficient to prove that a model has fully reproduced human aesthetic activity.

Xin Jin et al.'s deep drift-diffusion model is inspired by psychological models of decision processes. A convolutional network predicts drift-diffusion parameters, which are then used to predict aesthetic score distributions.[35] By using parameters of latent decision processes for score distribution prediction, the model connects an evaluation task with a process hypothesis. In the first instance, improvements in score distribution prediction support the model's effectiveness on that task. Whether its assumed accumulation process corresponds to actual human processes still requires independent evidence from response times, experimental manipulations, or neural observations.

### 5.2 The theoretical foundations of aesthetic process modeling: from emotional responses to cognitive mechanisms

Psychology and cognitive neuroscience have proposed several models of aesthetic processing. Leder et al. discuss perceptual analysis, memory integration, classification, and cognitive mastering; Pelowski et al. compare the structures and explanatory scopes of different models. [36, 37] Applying these theories to computational modeling requires defining the components and their variables, explaining how they correspond to observed data, and making testable predictions.

Emotion is one dimension of aesthetic experience. Studying aesthetic judgment also requires examining attentional selection, perceptual processing, memory associations, cultural background, and value judgment, using behavioral, physiological, and neural observations to analyze these processes.

Behavioral data include ratings, choices, and response times, which can record individuals' performance on specific tasks. Eye movements, fixations, and saccades can provide evidence of attentional allocation. Peripheral physiological indicators, such as heart rate, skin conductance, and respiration in Figure 5's “physiological/electrophysiological” category, can reflect aspects of response variation. Electroencephalography (EEG), magnetoencephalography (MEG), and functional magnetic resonance imaging (fMRI) provide neural evidence at different temporal and spatial scales. These data must be interpreted alongside experimental tasks, stimulus features, and individuals' reports. They neither correspond one-to-one with particular aesthetic processing components nor directly equate to aesthetic experience itself.

Researchers need to specify the hypothesized latent processes and their observable consequences, establish connections among stimuli, responses, and model variables, and use predictive testing, model comparison, and necessary experimental manipulations to determine whether a process hypothesis is preferable to alternative explanations. A well-fitted predictive relationship among observables is insufficient on its own to establish a latent-process hypothesis.

### 5.3 Computational neuroaesthetics and aesthetic judgment modeling

Computational neuroaesthetics seeks to connect stimulus representations, behavioral evaluations, and neural responses through models.[34] Using images of artworks, architecture, and natural landscapes, Vessel et al. compare within-domain and cross-domain information about aesthetic appeal through multivoxel fMRI pattern classification. They find that patterns in the default-mode network contain predictive information across visual domains.[38] This result supports representations of aesthetic evaluation across visual domains, but is insufficient to establish their causal role in the formation of judgments.

A further claim to explain mechanisms of aesthetic formation would require specifying how model variables relate to candidate neural processes and comparing alternative explanations.

Figure 5 is a conceptual framework that brings together aesthetic processing theories and computational research. Perceptual processing, attentional selection, affective response, memory and cognition, and value integration are functional components that can be analyzed. These processes may overlap and involve feedback, and are influenced by personal experience, cultural environment, social context, and current psychological state. The figure divides evidence into attention data, physiological/electrophysiological data, neural data, and behavioral data, with EEG included in neural data. The computational testing pathway identifies, in sequence, multimodal alignment, latent-process representation, mechanistic hypothesis fitting, and testing and revision. Its aim is to explain aesthetic judgment; testing still requires predictive testing, model comparison, and necessary experimental manipulations.

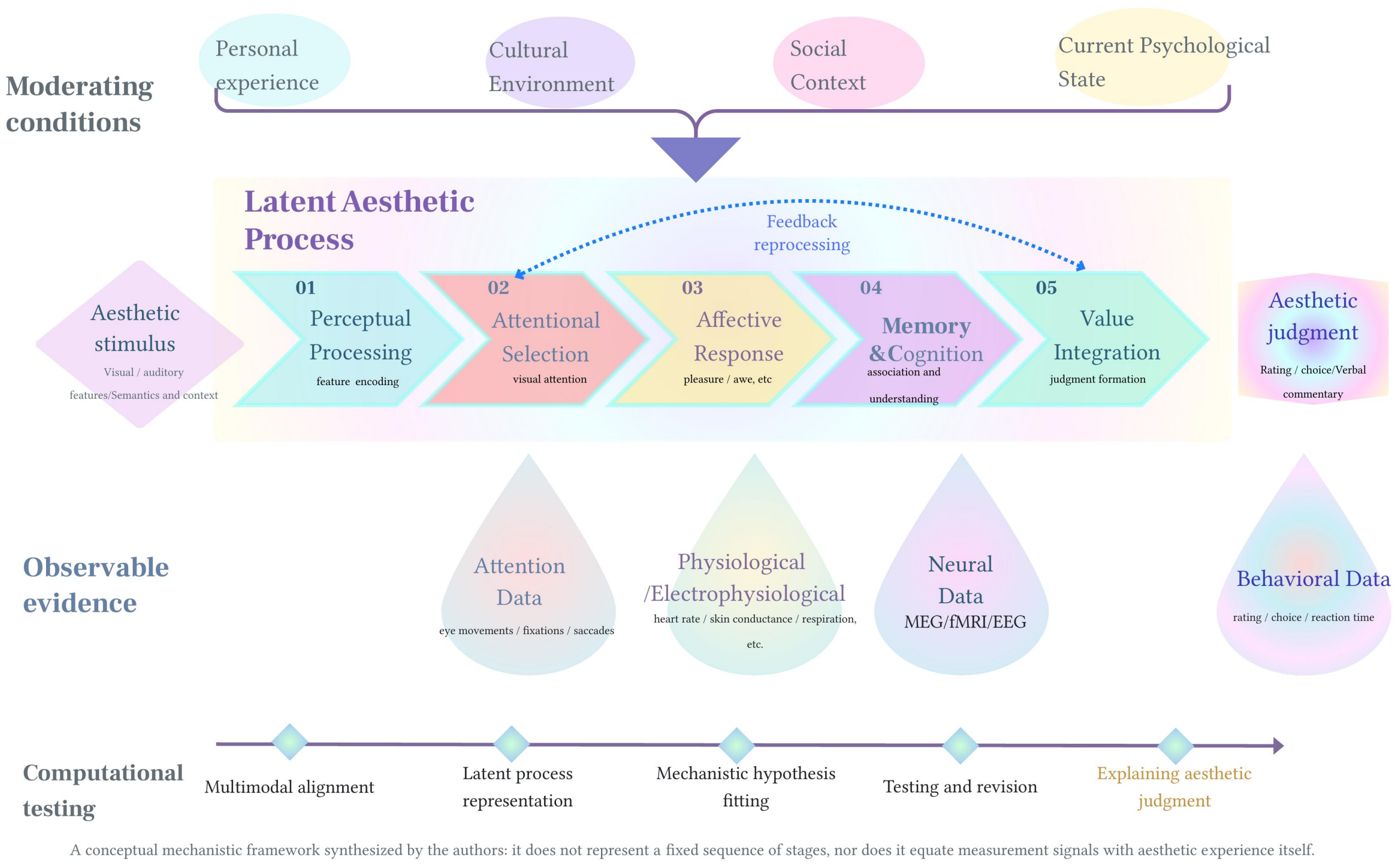

Figure 5. Latent Mechanisms, Observable Evidence, and Computational Testing in Aesthetic Process Modeling. Synthesized and drawn by the authors with reference to Leder et al. (2004), Pelowski et al. (2016), and Li and Zhang (2020). The arrangement of components and evidence and the testing pathway are our synthesis.

### 5.4 Aesthetic intelligence and future directions for computational aesthetics

Aesthetic measurement provides a computational basis for evaluation and selection, process modeling proposes testable hypotheses about how evaluations form, and generative systems can use these models to guide artistic creation. These three activities can provide feedback to one another within the same system or be pursued as relatively independent research questions. Integrated systems should report validation results for each capability separately.

Personalized and cross-media research especially requires a clearly specified evaluator. Group mean scores, individual preferences, and expert evaluations are not the same target. Photography, painting, video, and music also differ in how they are organized. Integrating aesthetic intelligence should therefore involve specifying the task, evaluator, and context of use, and separately testing evaluation prediction, process explanation, and assisted creation.

## 6 Discussion: Terminology Translation and Conceptual Usage

Cross-language accounts may amplify problems of conceptual boundaries. The English word aesthetic(s) can denote aesthetic qualities, theories and ideas about beauty or art, or aesthetics as a branch of philosophy.[39] In expressions such as "Chinese-style aesthetics," the Chinese term for "aesthetics" may also serve to describe a style or outlook. Distinguishing among these meanings requires attention to the original-language collocation, the object in question, and the context of discussion.

Table 2 compares the original-language expressions, the framing of the research question, and the relation to computational research across the relevant literature. AI as a research method differs from AI as an object of theoretical inquiry, just as studying the style of a work differs from using a term to name that style. AI Aesthetics and Aesthetic AI have different heads. These heads may suggest their expressive functions but are not by themselves sufficient to determine conceptual affiliation. Likewise, artificial in An Artificial Aesthetic cannot be equated with AI without explanation.

Table 2. Contexts and Classification of Aesthetic Terminology in the Relevant Literature.

| Work and Location | Original-Language Expression | Context of Discussion | Distinctions in we |
|---|---|---|---|
| Manovich 2018, pp. 3, 7 | Aesthetic AI; Measuring Diversity | Automation of aesthetic choices; formal measurement of aesthetic diversity | Its overall scope is broader; research using AI for scoring or formal measurement can be classified according to we's criteria |
| Arielli 2024, pp. 14–16 | object vs. subject; description vs. generation | Four application types based on object/subject and description/generation | Classifying applications is not equivalent to classifying tasks; simulating judgments does not automatically explain processes |
| Thon and Wilde 2025, pp. 2, 6, 9 | media aesthetics; artistics; aisthetics | Distinctions among AI media, artistic judgment, and aesthetic perception | A media-aesthetic discussion; cannot as a whole be treated as a subfield defined by AI methods |
| Boden and Edmonds 2019, p. 3 | An Artificial Aesthetic; computer art | Computer art and its relationship with traditional art | Discusses the theory and practice of computer art; artificial does not directly mean AI |
| Feng Tao 2018, newspaper p. 4 | "Its methods are primarily those of philosophical aesthetics" | AI-related sensibility, art, and theoretical questions | AI as a research object differs from AI as a computational method |

Cross-language accounts need to distinguish attributes of works from research fields, research objects from technical methods, and the prediction of evaluations from mechanistic explanation. Table 2 sets out the differences in questions and terminology among the works discussed; these differences do not in themselves show that any author has misused a concept. A critique of a particular translation requires comparing the original and translated texts, identifying where one meaning replaces another, and explaining how that substitution affects the argument. Uses in different disciplines should be read within their respective contexts.

## 7 Conclusion: The Special Core, General Boundaries, and Development of Computational Aesthetics

We distinguish between the special core and the general extension of computational aesthetics. The special core comprises aesthetic measurement and aesthetic process modeling, which address, respectively, the computational treatment of aesthetic attributes and evaluations and the modeling of how human aesthetic perception and judgment are formed. The general extension further encompasses artistic generation that is substantively guided by computational aesthetic models. AI aesthetics is the part of research within this scope that employs AI algorithms and models. It is defined by the technical methods it uses and does not constitute a separate, parallel category of task.

This distinction provides criteria for classifying the literature and for positioning research questions. AI aesthetics is a methodological subset of computational aesthetics; computational art, AIGC, computational creativity, and affective and neural research intersect with computational aesthetics conditionally, depending on the specific question at issue, or else supply relevant models, methods, and evidence. Feng Peng's discussion of the computability of beauty, aesthetic experience, and art offers a philosophical reference point for these questions. Assigning a study to a category requires considering its objects, objectives, methods, and evidence together, rather than relying solely on the name of a technology or on how a work was generated.

The prediction of aesthetic evaluations, model explainability, and explanations of human aesthetic mechanisms require separate testing. Research on generation must also explain how computational aesthetic models participate in artistic creation, while cross-language citation must preserve the original formulation of the research question and its context of use. The criteria we propose help to organize these distinctions; their applicability remains to be tested across different media, cultures, and individual subjects.